\pdfoutput=1
\documentclass[11pt]{article}

\usepackage[]{EMNLP2022}

\usepackage{times}
\usepackage{latexsym}

\usepackage[T1]{fontenc}
\usepackage[utf8]{inputenc}

\usepackage{microtype}

\usepackage{graphicx}
\usepackage{amsmath}
\usepackage{amsthm}
\usepackage{booktabs}
\usepackage[ruled,linesnumbered]{algorithm2e}
\usepackage{multirow}
\usepackage{amssymb}

\usepackage{multirow}
\usepackage{natbib}
\usepackage{array}
\usepackage{graphicx}
\usepackage{bbm}
\usepackage{float}
\usepackage{arydshln}
\usepackage{dsfont}
\usepackage[utf8]{inputenc}
\usepackage{cleveref}
\crefname{section}{§}{§§}
\Crefname{section}{§}{§§}
\usepackage{makecell}
\usepackage{bbding}
\usepackage{amssymb}% http://ctan.org/pkg/amssymb
\usepackage{pifont}% http://ctan.org/pkg/pifont
\title{Query-based Instance Discrimination Network \\
for Relational Triple Extraction}

\author{
\textbf{Zeqi Tan$^{1}$}, 
\textbf{Yongliang Shen$^{1}$},
\textbf{Xuming Hu$^{2}$},
\textbf{Wenqi Zhang$^{1}$},
\textbf{Xiaoxia Cheng$^{1}$},\\
\textbf{Weiming Lu$^{1}$\thanks{\textsuperscript{$\ast$} Corresponding author}},
\textbf{Yueting Zhuang$^{1}$}\\
 $^{1}$Zhejiang University, $^{2}$Tsinghua University\\
 $^{1}$\texttt{\{zqtan,syl,luwm,yzhuang\}@zju.edu.cn}\\
 $^{2}$\texttt{\{hxm19\}@mails.tsinghua.edu.cn}
 }

\begin{document}

\maketitle

\begin{abstract}

Joint entity and relation extraction has been a core task in the field of information extraction. Recent approaches usually consider the extraction of relational triples from a stereoscopic perspective, either learning a relation-specific tagger or separate classifiers for each relation type. However, they still suffer from error propagation, relation redundancy and lack of high-level connections between triples. To address these issues, we propose a novel query-based approach to construct instance-level representations for relational triples. By metric-based comparison between query embeddings and token embeddings, we can extract all types of triples in one step, thus eliminating the error propagation problem. In addition, we learn the instance-level representation of relational triples via contrastive learning. In this way, relational triples can not only enclose rich class-level semantics but also access to high-order global connections. Experimental results show that our proposed method achieves the state of the art on five widely used benchmarks.

\end{abstract}

\section{Introduction}

Extracting structured information from open domain texts is a long-standing study topic in NLP.
Joint entity and relation extraction aims to mine high-quality relational triples (\textit{subject, relation, object}) from unstructured texts. 
For example, in the sentence ``900 people cross the border between Malaysia and Singapore'', the subject entity \textit{900 people} and the object entity \textit{border} have a physical location relation \texttt{PHYS}.

The current entity and relation extraction methods can be divided into two categories: 
pipeline methods \citep{chan2011exploiting,lin2016neural} and joint methods \citep{miwa-bansal-2016-end,katiyar-cardie-2017-going}.
% The pipeline methods first use entity recognition models to label entities in a sentence (NER) and then predict the relations between them by another model (RE). 
% Despite the flexibility of pipeline methods, these methods face the error propagation problem, where prediction errors from entity recognition affect the subsequent relation extraction. In addition, the lack of effective interaction between the two tasks ignores the intrinsic connections and dependencies between them.
Despite the flexibility of pipeline methods, they face the error propagation and the lack of interaction problems \citep{katiyar-cardie-2017-going,wang2020two}.
Therefore, many strategies have been proposed to unify the two tasks. 
\citet{zheng2017joint} extends tagging schema to tag both entities and relations. However, it cannot handle the triple overlapping problem. 

\begin{figure}[t!]
  \centering
  \includegraphics[width=\linewidth]{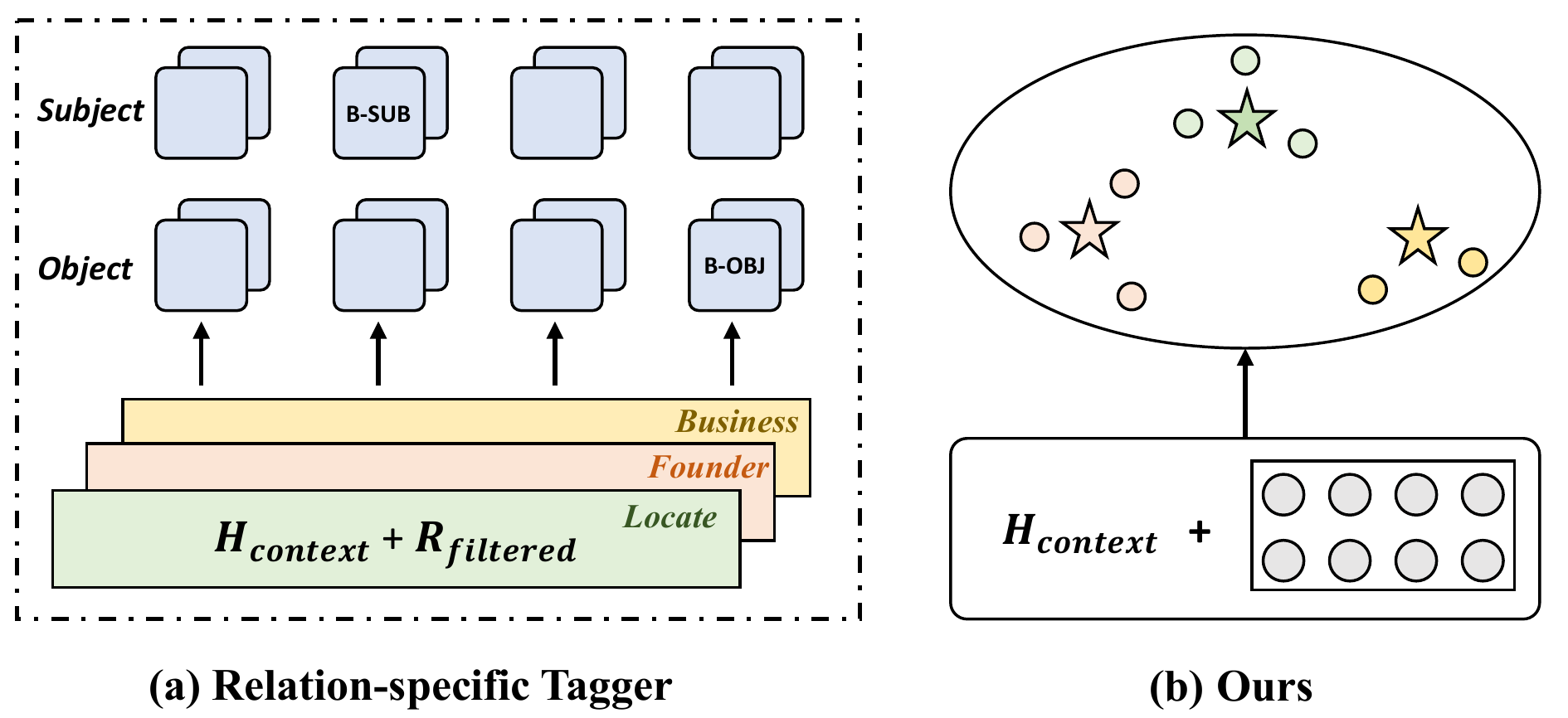}
  \caption{(a) $H_{context}$ represents sentence representation, $R_{potential}$ represents potential relation type embeddings. PRGC first filters out unlikely relations in a sentence, and then learns the relation-specific tagger for potential relations. (b) The gray circle below represents the initial state of query embeddings which are type-agnostic, and the star above represents relation type embeddings. We can break the limitation of type-independence and learn the high-order connections between different types via contrastive learning.}
  \label{fig:example}
\end{figure}

To tackle this problem, the generation-based methods \citep{zeng2018extracting,zeng2020copymtl,nayak2020effective,ye-2021-contrasive}, the table-filling methods \citep{wang-etal-2020-tplinker,wang2020two,wang-etal-2021-unire} and the cascading methods \citep{luan2019general,yuan2020,wei-etal-2020-novel,zheng-2021-prgc} have been investigated.

Recently, an increasing number of studies have come to consider the relational triple extraction task from a stereoscopic perspective. These approaches treat relation type as an important element, and either learn a distinct classifier \citep{liu2020,wang-etal-2020-tplinker} for each relation type or construct a relation-specific sequence tagger \citep{yuan2020,wei-etal-2020-novel,zheng-2021-prgc}.
In this way, these methods can better decouple the task-specific features for relations and entities, and yield promising results. However, they still have some weaknesses.
\textbf{First}, error propagation is a well-known problem. 
\citet{wei-etal-2020-novel} first identifies the subject and then extracts the corresponding triples. However, there are inevitable errors in the first stage. 
PRGC \citep{zheng-2021-prgc}, as shown in Figure \ref{fig:example} (a), first picks the potential relations in the sentence, and later tags the relation-specific entities. The correct relations are then likely to be filtered out in this process.
\textbf{Second}, these methods mostly suffer from the relation redundancy problem.
\citet{wei-etal-2020-novel,wang-etal-2020-tplinker} assembles triples on all relation types, which generates numerous invalid operations, leading to sparse label and low convergence rate.
\textbf{Third}, these methods process the triples for each relation type independently, ignoring the high-level global connections between different types of triples.
For example, if the triple (\textit{Maryland, country, U.S.}) has been recognized, then \textit{U.S}. is not supposed to appear in any other family relationship, such as \textit{spouse of}.

To tackle the above problems, we propose a novel query-based instance discrimination network (QIDN), in which we leverage instance query to construct instance-level representation for different types of relational triples. 
As in Figure \ref{fig:example} (b), we first use a set of type-agnostic query embeddings to obtain useful contextual information from a pre-trained language model. Afterwards, by the metric-based comparison between query embeddings and token embeddings, we can extract triples of all types in one step. 
In this manner, the error propagation problem is eliminated. Moreover, since we query the most plausible relation type directly with the query vector, the problem of relation redundancy is well mitigated.
In addition, we construct instance-level representations for relational triples by contrastive learning. 
Specifically, as shown in Figure \ref{fig:example} (b), we set two training objectives: (1) Intra-class instance pairs should get higher similarity than inter-class ones. (2) Instance representations should be closer to their corresponding relation type representations.
We aim to break the limitation of type-independence in previous methods and establish the global connection between triples.

Our main contributions are as follows:
\begin{itemize}
    \item We tackle the relational triple extraction task from a novel perspective, focusing on constructing instance-level representations for relational triples. Based on a simple similarity measure, we extract all types of triples in one step, thus avoiding cascading errors.
    \item We learn the instance-level representation of relational triples by contrastive learning. In this way, relational triples are not only able to establish high-order global connections, but also to enclose rich class-level semantics. 
    \item Extensive experiments on five public benchmarks demonstrate that our model achieves state-of-the-art, and significantly outperforms a range of robust baselines.
\end{itemize}

\section{Related Work}
\subsection{Mainstream methods}
Relational triple extraction can be solved with two different models. Assuming that the entities in a text sequence are obtained by NER model \citep{zhang-etal-2021-crowdsourcing,ijcai2021p0542,shen2022parallel,zhang-etal-2022-domain}, relation extraction can be considered as a classification task \citep{zeng-etal-2014-relation,wang-etal-2019-extracting,hu2020selfore,hu2021gradient}. Since these methods require additional entity annotators to carry out a pipeline, they generally face the error propagation problem and the lack of interaction between tasks \citep{lin2016neural,shen2021locate}. Recently, \citet{zhong-chen-2021-frustratingly}  propose a simple pipeline approach reaching state-of-the-art. They re-construct the input text of the relation model with the entity recognition results, delivering the type and location info of entities in this manner.

Beside the pipeline methods, the joint methods can be divided into the generation-based methods \citep{zeng2018extracting,zeng2020copymtl,nayak2020effective,ye-2021-contrasive}, the table-filling methods \citep{gupta2016table,zhang2017end,wang-etal-2020-tplinker,wang-etal-2021-unire} and the cascading methods \citep{luan2019general,yuan2020,wei-etal-2020-novel,zheng-2021-prgc}.
The generation-based methods use a sequence-to-sequence model to generate the relation triples directly from the sentence, while the table-filling methods treat on/off-diagonal entries as labels for entities and relationships, respectively. 
Differently, \citet{luan2019general} treats entity and relation extraction as a cascading two-level span classification task. Later, some work introduce a relation-specific perspective to treat triple extraction as a relation-first \citep{yuan2020,zheng-2021-prgc} or relation-middle \citep{wei-etal-2020-novel} cascading process. Compared to them, our method can form triples in one step and avoid the cascading errors.

\subsection{Query-based methods}
To exploit the well developed machine reading comprehension models, a number of works \citep{li2019entity,li-etal-2020-unified,ijcai2020-546,du-cardie-2020-event} extend the MRC paradigm to information extraction tasks.
\citet{li2019entity} recast the entity and relation extraction task as a multi-turn question answering problem.
Their queries are constructed based on pre-defined templates and the answers are the entity spans in the sentence. 
% After extracting the head entity, they construct the natural language queries based on the relation type to obtain the corresponding tail entity. 
% \citet{ijcai2020-546} considers that a single question template used in \citet{li2019entity} is not sufficient to express the diversity of contextual semantics. They attempt to cover rich relational semantics with more query templates based on subcategories of relations. 
\citet{ijcai2020-546} explores to cover more relational semantics with more handcrafted query templates based on subcategories of relations. 

Different from the above methods where queries are manually constructed, in the field of object detection in computer vision, there is a range of work \citep{carion2020end,zhu2020deformable,sun2020sparse} that uses trainable query to learn task-specific features automatically. 
DETR \citep{carion2020end} successfully integrates Transformer \citep{vaswani2017attention}, originally designed for text, by using a set of learnable queries. 
These queries are type-agnostic and have the potential to build global dependencies between objects at diverse scales. 
Inspired by this, we employ such query embeddings to construct instance-level representations for relational triples and establish global connections between different types of triples.

\subsection{Contrastive relation learning}
MTB \citep{soares2019matching} extracts relation-aware semantics from text by comparing sentences with the same entity pairs. Extended MTB, CP \citep{peng2020learning} samples relational triples with better diversity and increases the coverage of entity types and different contexts. Different from the sentence-level relational contrastive learning \citep{soares2019matching,peng2020learning,liu2022hierarchical}, ERICA \citep{qin2021erica} proposes a relation discrimination task to distinguish whether two relation types are semantically similar or not, better considering the interaction between multiple relations. 
Compared to them, we use query embeddings to construct instance-level triple representations, taking into account both entity features and relation features, which may motivate new pre-training tasks.

\section{Method}

\begin{figure*}[t!]
  \centering
  \includegraphics[width=\linewidth]{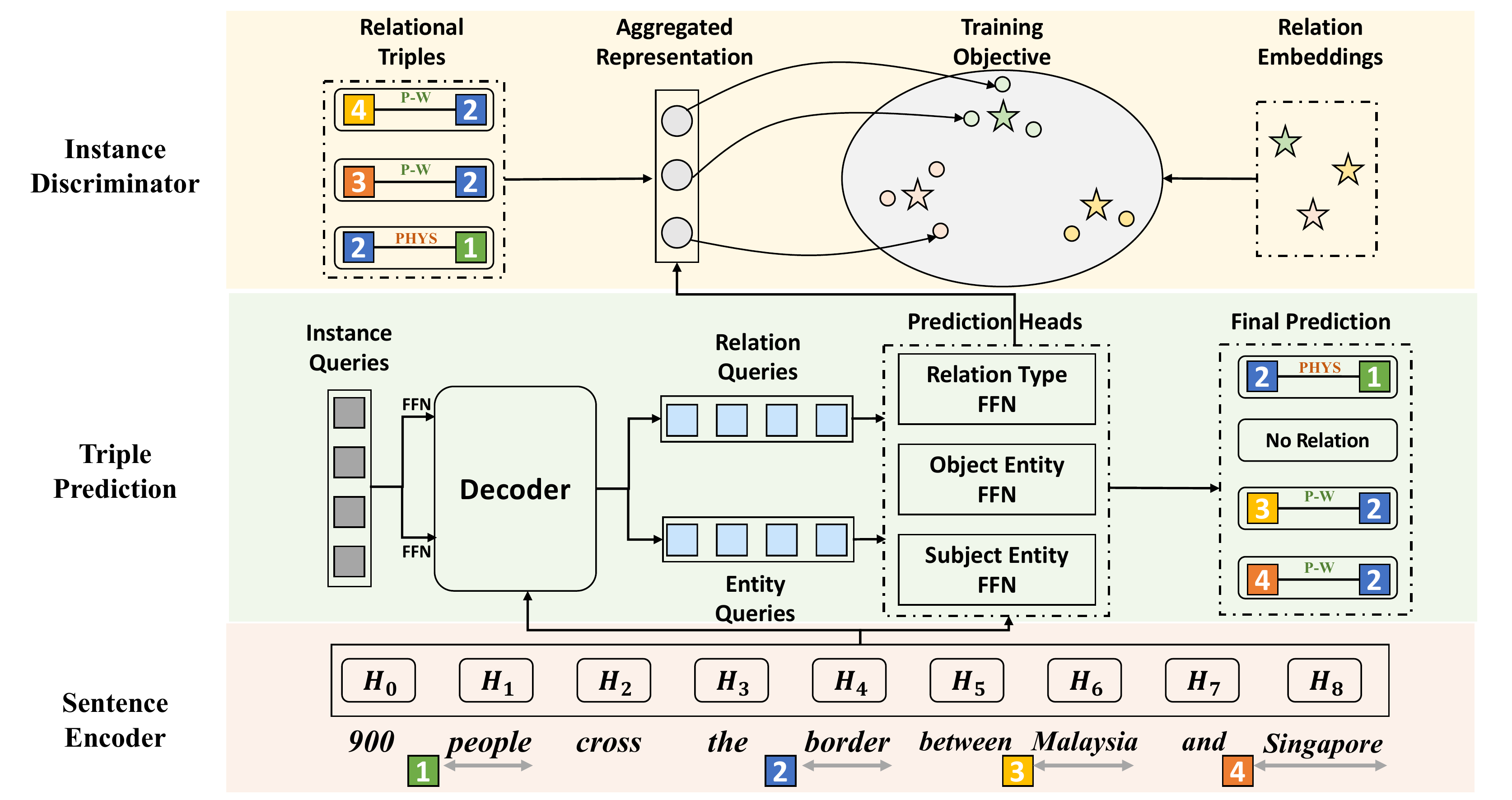}
  \caption{The overall architecture of our method. First we utilize the sentence encoder to get the context representation. Then, we employ a transformer-based decoder to transform a set of instance queries. Afterwards, we use several prediction heads to map the query vectors to different representation spaces for metric-based comparison with the token embeddings, so that all triples can be recognized in one step. Furthermore, we aggregate the representation of these prediction heads to construct instance-level representations for triples. By setting two contrastive learning training objectives, we can capture global connections between triples and rich class-level semantic information.}
  \label{fig:overview}
\end{figure*}
 
\subsection{Task Formulation}
\label{2.1}
The entity and relation extraction task aims to extract a set of entities and a set of relations from the text. Formally, given an input sentence $X = x_1, x_2, \ldots, x_{n}$ ($x_i$ is the i-th token, $n$ is the sentence length), an entity in the entity set $\mathcal{E}$ is denoted as $\left(x_{i}, x_{j}, t_{e}\right)$, where $x_i$ and $x_j$ are the left and right token of the entity with a pre-defined entity type $t_e$ in $\mathcal{Y}_e$. 
For the relation set $\mathcal{R}$, a relation is denoted as $\left(e_{1}, e_{2}, t_{r}\right)$, where $e_1, e_2 \in \mathcal{E}$ are the subject and object entities and $t_r$ is a pre-defined relation type in $\mathcal{Y}_r$. 
Besides, we add an additional label $\varnothing$ to $\mathcal{Y}_e$ and $\mathcal{Y}_r$ to indicate that no entity or no relation is recognized.

\subsection{Sentence Encoder}
\label{2.2}

As shown in Figure \ref{fig:overview}, given the input sentence $X$, we first use the pre-trained BERT model \citep{devlin-etal-2019-bert} to obtain a contextual representation for each token. 
Afterwards, to better consider the order information of the tokens, we feed the token representations into BiLSTM \citep{zhang-etal-2015-bidirectional} to get the final sentence representation $H \in \mathbb{R}^{n \times d}$, where $n$ is sentence length and $d$ is hidden size.

\subsection{Triple Prediction}
\label{2.3}
In our method, we employ a set of learnable instance queries same as DETR \citep{carion2020end}, which are denoted as $Q = \mathbb{R}^{M \times d}$.
Each query (denoted as a vector of size $d$) is responsible for extracting one relational triple.
These queries are randomly initialized and the number of the queries $M$ is pre-specified.  
Different from DETR, in order to distinguish entity and relation task-specific features, we project the queries $Q$ into entity and relation branches by FFN before entering the decoder.
After decoding, we use several prediction heads to map these query vectors to different representation spaces for metric-based comparison with the token embeddings.
In this way, all types of triples can be recognized in one step.

\paragraph{Transformer-based Decoder}
The decoder is composed of a stack of $L$ transformer \citep{vaswani2017attention} layers, and the decoding process mainly involves the multi-head attention mechanism. 
For simplicity, we denote the attention as:
\begin{equation}
\text { Attention }({Q}, {K}, {V})=\operatorname{softmax}\left(\frac{{Q K}^{T}}{\sqrt{d_{k}}}\right) {V},
\end{equation}
where ${Q}$, ${K}$, ${V}$ are the query, key and value matrix respectively, and the $1 / \sqrt{d_{k}}$ is the scaling factor. 
In the cross-attention mechanism module, we do not directly take $H$ from the sentence encoder as key and value, instead, we construct span-level representations to obtain hierarchical semantic information. 
Let $S = s_1, s_2, \ldots, s_{n_s}$ be all possible spans in sentence $X$ of up to limited length ($n_s$ is the number of spans). Give a span $s_{i} \in S$, the span representation $H_{\mathrm{i}}^{span}$ is defined as:
\begin{equation}
H_{\mathrm{i}}^{span}=\left[H_{\mathrm{start}(\mathrm{i})} ; H_{\operatorname{end}(\mathrm{i})} ; \phi\left(s_{\mathrm{i}}\right)\right],
\end{equation}
where $[;]$ denotes concatenation operation, $H_{\mathrm{start}(\mathrm{i})}$ and $H_{\mathrm{end}(\mathrm{i})}$ are the representations of the boundary tokens. $\phi\left(s_{\mathrm{i}}\right)$ denotes the span-length feature \citep{zhong-chen-2021-frustratingly}. Then, the span-level representations is denoted as $H^{span} \in \mathbb{R}^{n_s \times \mathrm{d}}$.
Through the $L$ decoding layers, instance queries $Q$ are decoded into $Q_r$ and $Q_e$ as:
\begin{equation}
[Q_r;Q_e] =\text {Decoder}\left([Q W_r;Q W_e], H^{span}\right),
\end{equation}
where $W_r, W_e\in \mathbb{R}^{d \times d}$ are trainable parameters, $Q_r, Q_e \in \mathbb{R}^{M \times d}$ denote relation and entity queries.
% Please refer to Appendix \ref{app:analysis} for the architecture illustration of the decoder.
For a more intuitive understanding, we illustrate the architecture of the decoder in Appendix \ref{app:analysis}.

\paragraph{Relation Head}
For relation queries $Q_{r}$, we feed them into a single layer of FFN to predict the categories of their corresponding triples.
Formally, we define the probability of the $i$-th query belonging to type $c$ as:
\begin{equation}
    P^t_{ic} = \frac{\exp(Q_r^iW^c_t  + b_t^c)}{\sum_{c^{\prime}}^{|\mathcal{Y}_r|}\exp(Q_r^iW^{c^{\prime}}_t + b_t^{c^{\prime}})},
\end{equation}
where $W_t\in \mathbb{R}^{|\mathcal{Y}_r| \times d}$ and $b_t \in \mathbb{R}^{|\mathcal{Y}_r|}$ are trainable parameters.

\paragraph{Entity Head}
To predict the boundary tokens of triples, we first perform linear projections for the entity queries $Q_e$ and the token representations $H$ by a single layer of FFN as:
\begin{equation}
E_\delta=Q_e W_\delta, H_{s}=HW_s,
\end{equation}
where $\delta \in \mathcal{C} = \{l_{sub}, r_{sub}, l_{obj}, r_{obj}\}$ denotes the left or right boundaries of subject or object entities and $W_{\delta}, W_s \in \mathbb{R}^{d \times d}$ are projection parameters.
To measure the similarity between them, we adopt cosine similarity function $S(\cdot)$ as:
\begin{equation}\label{sim}
S\left(\mathbf{v}_{i}, \mathbf{v}_{j}\right)=\frac{\mathbf{v}_{i}}{\left\|\mathbf{v}_{i}\right\|} \cdot \frac{\mathbf{v}_{j}}{\left\|\mathbf{v}_{j}\right\|}.
\end{equation}

Then, we can calculate the probability that the $j$-th token is the boundary token of the $i$-th entity queries as: 
\begin{equation}
    P^\delta_{ij} = \frac{\exp S\left(E_\delta^i, H_s^j\right)}{\sum_{j^{\prime}}^{n}\exp  S\left(E_\delta^i, H_s^{j^{\prime}}\right)},
\end{equation}
where $n$ is the number of tokens in the sentence. Finally, based on the probability $P^t$ and $P^\delta$ ($\delta \in \mathcal{C}$), we can predict all types of triples. Besides, for corpus with labeled entity types, we add two additional entity type heads like the relation head.

\subsection{Instance Discriminator}
In this module, we first aggregate several representations from the prediction heads to form the initial representation of triples, and then we set up two contrastive learning objectives to build global connections between triple instances and learn rich class-level semantic information.

\paragraph{Aggregation}
To get the initial representation of triples, we first use FFN to project the relation queries $Q_r$, aligned with $E_\delta$ from the entity head. Then, we aggregate category representations and boundary representations by a simple summing operation. The triple instance representation $\mathbf{v}$ is:
\begin{equation}
\mathbf{v} = Q_rW + \sum_{\delta \in \mathcal{C}} E_\delta.
\end{equation}
% where $W\in\mathbb{R}^{d \times d}$ are trainable parameters.

\paragraph{Training Objective}
With the triple instance representation $\mathbf{v}$ and the similarity function defined in equation \ref{sim}, we discuss how to set our training objectives.
Denote $\mathcal{R}=\left\{\mathbf{r}_{1}, \cdots, \mathbf{r}_{|\mathcal{Y}_r|}\right\}$ as the set of relation type embeddings which are randomly initialized.
We expect to achieve two goals with optimization: (1) For instance-instance pairs, intra-class pairs should get higher similarity than inter-class ones. (2) For instance-type pairs, instance representations should be closer to their corresponding relation type representations.
To reach these two goals, we define the objective function based on InfoNCE \citep{DBLP:journals/corr/abs-1807-03748}, which is widely applied in contrastive learning.

For the instance-instance objective, we minimize the following loss:
\begin{equation}
\mathcal{L}_{\text{ins}}= - \sum_{c} \sum_{i, j} \log \frac{\exp S\left(\mathbf{v}_{i}^{c}, \mathbf{v}_{j}^{c}\right)}{\sum_{c \prime, j^{\prime}} \exp S\left(\mathbf{v}_{i}^{c}, \mathbf{v}_{j^{\prime}}^{c \prime}\right)},
\end{equation}
where $(\mathbf{v}_{i}^{c}, \mathbf{v}_{j}^{c})$ denotes the instance pair of the same type $c$. 
% We aim to break the limitation of type-independence and establish global connection between triples.
Similarly, the instance-type loss is defined as:
\begin{equation}
\mathcal{L}_{\text{cls}}= - \sum_{i, c} \log \frac{\exp S\left(\mathbf{v}_{i}^{c}, \mathbf{r}_{c}\right)}{\sum_{c \prime} \exp S\left(\mathbf{v}_{i}^{c}, \mathbf{r}_{c^{\prime}}\right)},
\end{equation}
where $\mathbf{v}_{i}^{c}$ is the instance of type $c$ and $\mathbf{r}_{c} \in \mathcal{R}$ is the relation embedding corresponding to type $c$. 
% This objective allow triples to enclose rich class-level semantic information.
We aim to break the limitation of type-independence and establish the global connection between different types of triples.
% For the instance-instance objective, we minimize the following loss:
% \begin{equation}
% \mathcal{L}_{\text{ins}}= - \sum_{i, j}  \mathds{1}_{y_i=y_j} \log \frac{\exp S\left(\mathbf{v}_{i}, \mathbf{v}_{j}\right)}{\sum_{j^{\prime}} \exp S\left(\mathbf{v}_{i}, \mathbf{v}_{j^{\prime}}\right)},
% \end{equation}
% where $y_i$ denotes the type of the instance $i$. 
% % We aim to break the limitation of type-independence and establish global connection between triples.
% Similarly, the instance-type loss is defined as:
% \begin{equation}
% \mathcal{L}_{\text{cls}}= - \sum_{i, c} \mathds{1}_{y_i=c} \log \frac{ \exp S\left(\mathbf{v}_{i}, \mathbf{r}_{c}\right)}{\sum_{c \prime} \exp S\left(\mathbf{v}_{i}, \mathbf{r}_{c^{\prime}}\right)},
% \end{equation}
% where $\mathbf{r}_{c} \in \mathcal{R}$ is the relation embedding corresponding to type $c$. 
% % This objective allow triples to enclose rich class-level semantic information.
% We aim to break the limitation of type-independence and establish global connection between different types of triples.

% Table generated by Excel2LaTeX from sheet 'Sheet1'
\begin{table}[t!]
\centering
\small
	\renewcommand\arraystretch{1.0}
    \begin{tabular}{lcccc}
    \toprule
    \multirow{2}{*}{\textbf{Dataset}} & \multicolumn{3}{c}{\textbf{\#Sentences}} & \multirow{2}{*}{\textbf{Relations}} \\
    \cmidrule(l){2-4}
     & \textbf{Train} & \textbf{Dev} & \textbf{Test} & \\
    \midrule
    NYT & 56,195 & 4,999 & 5,000 & 24 \\
    WebNLG & 5,019 & 500 & 703 & 170 \\
    NYT$^*$ & 56,195 & 5,000 & 5,000 & 24 \\
    ACE05 & 10,051 & 2,424 & 2,050 & 6 \\
    SciERC & 1,861 & 275 & 551  &7 \\
    \bottomrule
    \end{tabular}%
  \caption{Dataset statistics.}
  \label{tab:statistics}%
\end{table}%

\subsection{Training and Inference} 

\paragraph{Training}
In training, we define the triple loss $\mathcal{L}_{tri}$ according to the type probability $P^t$ and the boundary probability $P^\delta$ as:
\begin{equation}
\mathcal{L}_{\text{tri}} = - \sum_{i=1}^{M} \left(\log P^t_{\sigma(i)} + \sum_{\delta \in \mathcal{C}} \log P^\delta_{\sigma(i)} \right),
\end{equation}
where $M$ is the number of instance queries, $\sigma$ is the optimum matching calculated as same as \citet{carion2020end}.
The final loss function $\mathcal{L}$ is computed as $\mathcal{L} = \mathcal{L}_{tri} + \mathcal{L}_{ins} + \mathcal{L}_{cls}$.

\paragraph{Inference}
 During inference, the triple predicted by the $i$-th instance query is $\mathcal{Y}_i = \left( \mathcal{Y}^t_i, \mathcal{Y}^\delta_i\right)$, $\delta \in \mathcal{C}$.
 $\mathcal{Y}^{t}_i = \mathop{\arg\max}_c(P^t_{ic})$ is the relation type.
 $\mathcal{Y}^\delta_i = \mathop{\arg\max}_k(  P^\delta_{ik})$ are the left and right boundary tokens of subject-object entities, and triples with the predicted type of $\varnothing$ will be filtered out.

\begin{table*}[t!]
	\centering
	\small
	\setlength{\tabcolsep}{3.2mm}
	\renewcommand\arraystretch{1.0}
	\scalebox{1}{
			\begin{tabular}{@{}lccccccccc@{}}
				\toprule
				\multicolumn{1}{c}{\multirow{2}{*}{Model}}                      & \multicolumn{3}{c}{NYT} & \multicolumn{3}{c}{WebNLG} & \multicolumn{3}{c}{NYT$^*$} \\ 
				\cmidrule(l){2-4} \cmidrule(l){5-7} \cmidrule(l){8-10} 
		        &   Prec.   & Rec.  & F1    & Prec.    & Rec.   & F1     & Prec.  & Rec.   & F1    \\  
				\midrule
				GraphRel \cite{fu-etal-2019-graphrel}         & 63.9    & 60.0  & 61.9  & 44.7     & 41.1   & 42.9   & -      & -      & -    \\
				RSAN \cite{yuan2020}            & -       & -     & -     & -        & -      & -      & 85.7  & 83.6  & 84.6 \\ 
				MHSA  \cite{liu2020}  & 88.1   & 78.5 & 83.0 &  89.5 & 86.0 & 87.7 & -   & -   & - \\
				\midrule[0.1pt]
				CasRel \cite{wei-etal-2020-novel}          & 89.7    & 89.5  & 89.6  & 93.4     & 90.1   & 91.8   & -      & -      & -     \\
				TPLinker \cite{wang-etal-2020-tplinker}        & 91.3    & 92.5  & 91.9  & 91.8     & 92.0   & 91.9   & 91.4   & 92.6   & 92.0  \\
				SPN \cite{sui2020joint}           & 93.3    & 91.7  & 92.5  & 93.1     & 93.6   & 93.4   & 92.5   & 92.2   & 92.3  \\
				CGT  \cite{ye-2021-contrasive}            & \textbf{94.7}    & 84.2  & 89.1  & 92.9     & 75.6   & 83.4   & -      & -      & - \\
				CasDE \cite{ma-dual-2021}    & 90.2    & 90.9  & 90.5  & 90.3     & 91.5   & 90.9   & 89.9   & 91.4   & 90.6  \\
				RIFRE  \cite{zhao-2021-kbs}         & 93.6    & 90.5  & 92.0  & 93.3     & 92.0   & 92.6   & -      & -      & -  \\
				PRGC  \cite{zheng-2021-prgc}           & 93.3    & 91.9  & 92.6  & 94.0     & 92.1   & 93.0   & \textbf{93.5}   & 91.9   & 92.7 \\ 
				\midrule
				QIDN           & 93.4    & \textbf{92.6}  & \textbf{93.0}  & \textbf{94.1}     & \textbf{93.7}   & \textbf{93.9}   & 93.3  &  \textbf{92.5}  & \textbf{92.9}  \\ 
				\bottomrule
			\end{tabular}
	}
	\caption{Precision(\%), Recall (\%) and F1-score (\%) of our method and baselines. Note that the sentence encoders adopted in GraphRel, RSAN and MHSA are LSTM networks, while other baselines employ the BERT model.}
	\label{tab:main}
\end{table*}

\section{Experiments}

\subsection{Experimental Setup}
\paragraph{Datasets}
We conduct our experiments on four widely used datasets: 
NYT \citep{riedel2010}, 
WebNLG \citep{zeng2018extracting},
ACE05 \citep{2005-automatic},
and SciERC \citep{luan-etal-2018-multi}.
NYT \citep{riedel2010} is sampled from New York Times news articles and annotated by distant supervision. 
NYT$^*$ is another version of it which annotates the whole span of entities.
WebNLG is originally created for Natural Language Generation (NLG) and is adopted by \citep{zeng2018extracting} as a relation extraction dataset. 
ACE05 corpora are collected from a wide range of domains, such as newswire and online forums.
SciERC includes annotations for scientific entities, their relations, and coreference clusters for 500 scientific abstracts. 
Table \ref{tab:statistics} shows the detailed dataset statistics. 

% \subsection{Evaluation Metrics}
\paragraph{Evaluation Metrics}
Following previous works \citep{wei-etal-2020-novel,zheng-2021-prgc}, we adopt micro Precision (Prec.), Recall (Rec.) and F1-score (F1) on NYT/WebNLG under partial match, on NYT$^*$ under exact match. For ACE05 and SciERC, same as \citep{zhong-chen-2021-frustratingly,yan-etal-2021-partition}, we adopt the micro F1-score as the evaluation metric for both NER and RE. We use the strict evaluation criterion that a relation is correct only if the relation type is correct and the type as well as boundaries of its corresponding subject-object entities are correct. 

% \subsection{Implementation Details}
\paragraph{Implementation Details}
For fair comparison with prior work, we use
\textit{bert-base-cased} \citep{devlin-etal-2019-bert} for NYT/WebNLG,
\textit{albert-xxlarge-v1} \citep{Lan2020ALBERT} for ACE05, and \textit{scibert-scivocab-uncased} \citep{beltagy-etal-2019-scibert} for SciERC in the sentence encoder. 
The instance queries and relation embeddings are randomly initialized with the normal distribution $\mathcal{N}(0.0, 0.02)$ and we design a comparison experiment in Appendix \ref{app:queries} to determine the query number $M$ as 15. 
% Please refer to Appendix \ref{app:queries} for more details.
The number of the BiLSTM layer is 3 and the number of our decoder layer $L$ is set to 5. The span enumeration length is set to 8. We adopt a batch size of 8 on NYT and ACE05, and 4 on WebNLG and SciERC in training. We use the AdamW \citep{DBLP:journals/corr/abs-1711-05101} optimizer with a linear warmup-decay learning rate schedule. The peak learning rate is set to 1e-5 for the pre-trained model and 3e-5 for the other parameters. We train each model for 100 epochs with a dropout of 0.1 \citep{srivastava2014dropout}.
% And all our experiments are conducted on a single NVIDIA RTX A6000 GPU.

\begin{table}[t!]
\centering
\renewcommand\arraystretch{1.0}
\small
\begin{tabular}{lcc}
\toprule
Model & NER & RE\\
\midrule
\multicolumn{3}{l}{\textbf{ACE05}} \\
Structured Perceptron \citep{li2014incremental} & 80.8 & 49.5 \\ 
SPTree \citep{miwa-bansal-2016-end} & 83.4 & 55.6 \\
Multi-turn QA \citep{li2019entity} $^{\dag}$ & 84.8 & 60.2 \\
Table-Sequence \citep{wang2020two} $^{\ddag}$ & 89.5 & 64.3 \\
Trigger-Sense \citep{shen2021trigger} $^{\dag}$ & 87.6 &  62.8  \\
PURE \citep{zhong-chen-2021-frustratingly} $^{\ddag}$  & 89.7 & 65.6  \\
UniRE \citep{wang-etal-2021-unire} $^{\ddag}$  & 90.2  & 66.0 \\
PFN  \citep{yan-etal-2021-partition} $^{\ddag}$  & 89.0  & 66.8 \\
\midrule
QIDN $^{\ddag}$  & \textbf{90.5} & \textbf{68.2}  \\
\midrule
\multicolumn{3}{l}{\textbf{SciERC}} \\
% \midrule
SPE \citep{wang-etal-2020-pre} $^{\S}$  & 68.0 & 34.6  \\
PURE \citep{zhong-chen-2021-frustratingly} $^{\S}$  & 66.6 & 35.6  \\
UniRE \citep{wang-etal-2021-unire} $^{\S}$  & 68.4 & 36.9  \\
PFN \citep{yan-etal-2021-partition} $^{\S}$  & 66.8  & 38.4 \\
\midrule
QIDN $^{\S}$ & \textbf{69.8}  & \textbf{39.5} \\
\bottomrule
\end{tabular}
\caption{The overall performances of our method on ACE05 and SciERC. 
$^{\dag}$, $^{\ddag}$ and $^{\S}$ denotes the use of BERT, ALBERT and SciBERT\citep{devlin-etal-2019-bert,Lan2020ALBERT,beltagy-etal-2019-scibert} pre-trained language models.}
\label{tab:overall}
\end{table}

\begin{table*}[t!]
% \small
	\setlength\tabcolsep{1.2mm}
	\renewcommand\arraystretch{1.0}
	\centering
	\scalebox{0.8}{
		\begin{tabular}{@{}lcccccccccccccccccc@{}}
			\toprule
			\multicolumn{1}{c}{\multirow{2}{*}{Model}} & \multicolumn{9}{c}{NYT}                                                   & \multicolumn{9}{c}{WebNLG}                                                \\ 
			\cmidrule(l){2-10} \cmidrule(l){11-19} 
			\multicolumn{1}{c}{}                       & Normal & EPO  & SEO  & SOO & N=1  & N=2  & N=3  & N=4  & N$\geq$5 & Normal & EPO  & SEO  & SOO & N=1  & N=2  & N=3  & N=4  & N$\geq$5 \\ 
			\midrule
			CasRel                                     & 87.3   & 92.0 & 91.4 &  77.0$^\S$   & 88.2 & 90.3 & 91.9 & 94.2 & 83.7         & 89.4   & 94.7 & 92.2 &  90.4$^\S$ & 89.3 & 90.8 & 94.2 & 92.4 & 90.9             \\
			TPLinker                                   & 90.1   & 94.0 & 93.4 & 90.1$^\S$  & 90.0 & 92.8 & 93.1 & 96.1 & 90.0             & 87.9   & 95.3 & 92.5 &  86.0$^\S$ & 88.0 & 90.1 & 94.6 & 93.3 & 91.6             \\
			SPN                                      & 90.8   & 94.1 & 94.0 &  -  & 90.9 & 93.4 & \textbf{94.2} & 95.5 & 90.6                & -     &  -   &    -  &   - & 89.5 & 91.3 & \textbf{96.4} & 94.7 & 93.8             \\
			PRGC                                       & \textbf{91.0}   & 94.5 & 94.0 &  81.8  & \textbf{91.1} & 93.0 & 93.5 & 95.5 & 93.0           & 90.4   & \textbf{95.9} & 93.6 &  94.6 & 89.9 & 91.6 & 95.0 & 94.8 & 92.8             \\
			\midrule
			QIDN                                   & \textbf{91.2}   & \textbf{94.9}  & \textbf{94.8} &   \textbf{90.7}  & 90.6 & \textbf{93.6} & 94.1 & \textbf{95.8} & \textbf{94.3}         &  \textbf{91.5} & 95.4  & \textbf{94.8}  & \textbf{94.9} & \textbf{91.2} & \textbf{92.8}   & 96.1 & \textbf{95.4} & \textbf{94.2}                \\ 
			\bottomrule
		\end{tabular}
	}
	\caption{F1-measure (\%) on sentences with different overlapping patterns and different triple numbers. $\S$ marks the results reported by \cite{zheng-2021-prgc}.}
	\label{tab:type}
\end{table*}

\subsection{Overall Performance}

Table \ref{tab:main} shows the overall performance of our proposed method (QIDN) as well as the baseline models on the NYT and WebNLG datasets. Overall, our method outperforms all the baseline model consistently, and achieves the new state-of-the-art. 
Compared with the robust baseline models CasRel \citep{wei-etal-2020-novel} and TPLinker \citep{wang-etal-2020-tplinker}, our model makes a significant improvement of +2.1\% and +2.0\% in absolute F1-measure on WebNLG.
Compared to the best cascade method PRGC \citep{zheng-2021-prgc}, our method achieves consistent improvement in terms of all three evaluation metrics on WebNLG.
Similarly, Table \ref{tab:overall} indicates that our method reaches state-of-the-art on both the NER and RE tasks for ACE05 and SciERC.
In comparison to the best joint method \citep{yan-etal-2021-partition}, we achieve superior performance in F1-measure on ACE05 and SciERC (+1.5\% and +3.0\% for NER, +1.4\% and +1.1\% for RE). 
The experimental results demonstrate the effectiveness of our proposed method on the joint entity and relation extraction task.
We believe the main improvement comes from the fact that our approach avoids cascading errors and the relation redundancy problem. The cascading methods \citep{wei-etal-2020-novel,zheng-2021-prgc} inevitably suffer from cascading errors. And learning separate classifiers for each relation type \citep{wang-etal-2020-tplinker,wang-etal-2021-unire} causes the problem of relation redundancy. Our approach solves these problems simultaneously and thus achieves promising results.

\subsection{Analysis on Complex Scenarios}
Following \citet{wei-etal-2020-novel,wang-etal-2020-tplinker,zheng-2021-prgc}, we evaluate our model on different triple overlapping patterns and different triple numbers.
The triple overlapping problem refers to triples sharing the same single entity (SEO, i.e. SingleEntityOverlap) or entity pair (EPO, i.e. EntityPairOverlap). 
For example, In ``Alice and Joe were born in the US'', triples (\textit{Alice, birthplace, USA}) and (\textit{Joe, birthplace, USA}) share the single entity \textit{USA}, while triples (\textit{Alice, birthplace, USA}) and (\textit{Alice, residence, USA}) share the whole pair. The detailed statistic is described in Appendix.

As shown in Table \ref{tab:type}, we achieve the best performance on all three overlapping patterns on WebNLG and NYT. 
In the normal set, we achieve +1.1\% improvement in F1-measure on WebNLG, while on NYT the improvement is very slight. We argue that this is because NYT is generated with distant supervision, and annotations are often incomplete, especially for the normal pattern.
In addition, as in Table \ref{tab:type}, our model performs well for both datasets with different number of triples, especially for the sentences with more than 5 triples (+1.3\% on NYT and +1.4\% on WebNLG). 
Compared to previous methods, our approach can break the limitation of type-independence and establish the global connection between different triples, and thus be more robust than baselines when dealing with these challenging complex scenarios.

% Table generated by Excel2LaTeX from sheet 'Sheet4'
\begin{table}[t!]
  \centering
  \small
  \setlength\tabcolsep{1.8mm}
  \renewcommand\arraystretch{1.0}
    \begin{tabular}{llllccc}
    \toprule
    \multirow{2}[0]{*}{Model} & \multicolumn{3}{c}{NYT} & \multicolumn{3}{c}{WebNLG} \\
\cmidrule(lr){2-4} \cmidrule(lr){5-7}          & \multicolumn{1}{c}{P} & \multicolumn{1}{c}{R} & \multicolumn{1}{c}{F} & P     & R     & F \\
    \midrule
    Default & \textbf{93.4}    & \textbf{92.6}  & \textbf{93.0}  & \textbf{94.1}     & \textbf{93.7}   & \textbf{93.9}  \\
    \midrule
    w/o $H_{span}$ & 92.9  & 92.6  & 92.8  & 93.5  & 93.0  & 93.3  \\
    w/o $Q_e, Q_r$ & 93.2  & 92.2  & 92.7  & 93.4  & 92.8  & 93.1  \\
    w/o $\mathcal{L}_{\text{ins}}$ & 92.0  & 92.5  & 92.2  & 93.0  & 92.8  & 92.9  \\
    w/o $\mathcal{L}_{\text{cls}}$ & 92.7  & 92.2  & 92.4  & 93.7  & 92.5  & 93.1  \\
    w/o $\mathcal{L}_{\text{ins}}, \mathcal{L}_{\text{cls}}$ & 91.1  & 92.6  & 91.8  & 93.3  & 91.8  & 92.5  \\
    \bottomrule
    \end{tabular}%
  \caption{Ablation studies with five different settings of our model on NYT and WebNLG.}
  \label{tab:interaction}%
\end{table}%

\subsection{Ablation Study}
\label{ab}
In this section, we take a closer look at the modules in our model that contribute to performance with five settings.
(1) \textbf{w/o $H_{span}$}: replace the span-level representation with the token-level representation,
(2) \textbf{w/o $Q_e, Q_r$}: remove entity and relation query branches and use only instance queries,
(3) \textbf{w/o $\mathcal{L}_{\text{ins}}$}: remove the training objective between instance pairs,
(4) \textbf{w/o $\mathcal{L}_{\text{cls}}$}: remove the training objective between instances and relation embeddings,
(5) \textbf{w/o $\mathcal{L}_{\text{ins}}, \mathcal{L}_{\text{cls}}$}: remove both contrastive learning training objectives.

From Table \ref{tab:interaction} we observe that \textbf{w/o $\mathcal{L}_{\text{ins}}, \mathcal{L}_{\text{cls}}$} leads to the most significant performance decrease in absolute F1-mearsure (-1.2\% on NYT and -1.4\% on WebNLG) and removing any of them causes visible performance drops, which demonstrates the effectiveness of the two contrastive learning objectives we set. In addition, as show in Table \ref{tab:interaction}, constructing span-level sentence representations brings performance gains (+0.2\% on NYT and +0.6\% on WebNLG), which indicates that span-level representations contain structured information that is more useful for triple extraction than token-level representations. Similarly, dividing the instance queries into entity and relation branches to facilitate the distinction between their task-specific characteristics, delivering performance improvement.
We provide a more detailed analysis for entity and relation branches in Appendix \ref{app:analysis}.

\begin{figure}[h!]
  \centering
  \includegraphics[width=\linewidth]{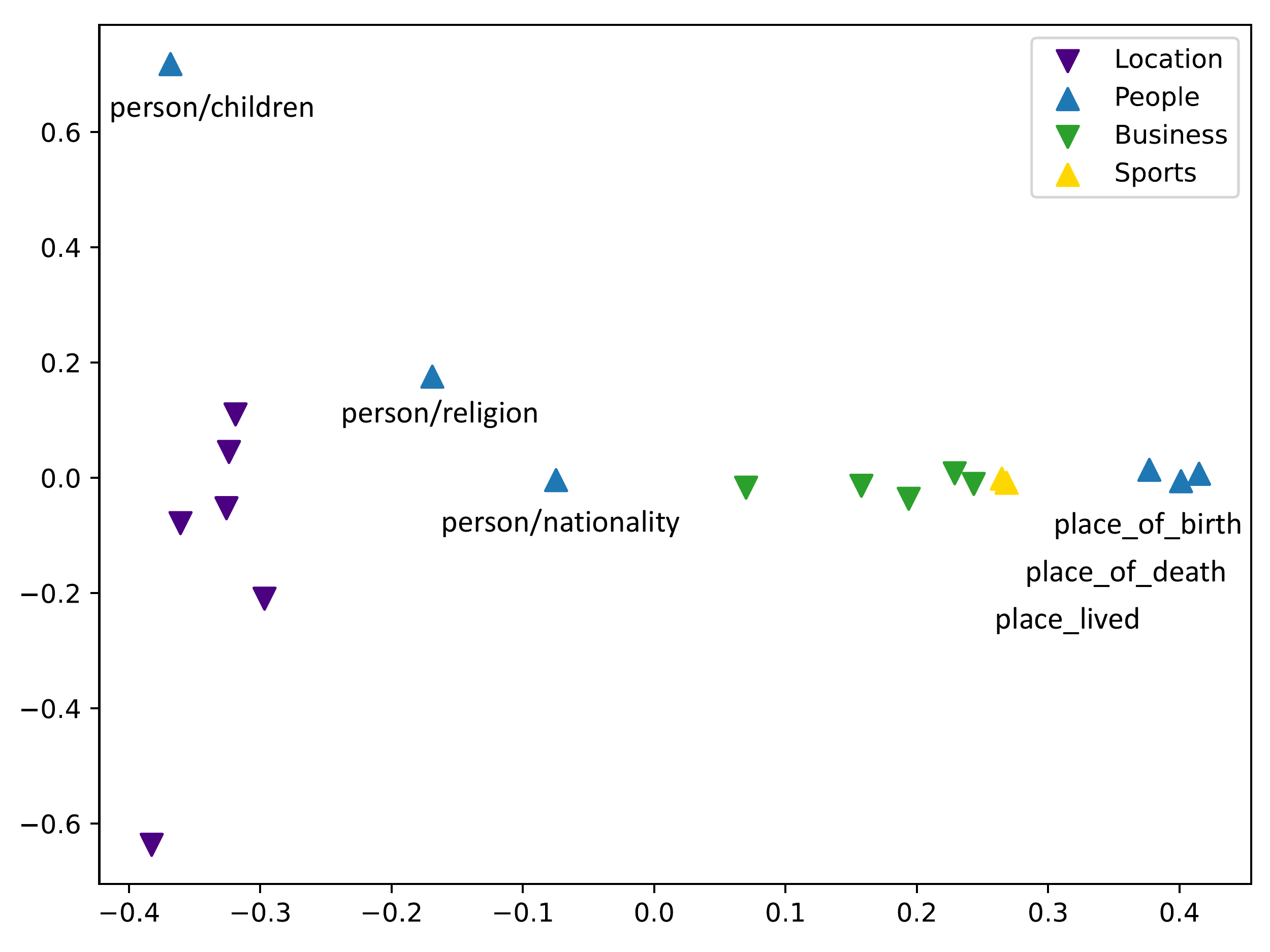}
  \caption{Visualization of relations on NYT dataset. For simplicity, we only label the fine-grained categories under ``People''.}
  \label{fig:cluster}
\end{figure}

\section{Topology of Relations}
To illustrate that our approach establishes connections between different types of relations, we visualize the relation representation learned on NYT.
We first perform L2 normalization on the relation embeddings and then use PCA \citep{abdi2010principal} to reduce their dimension.
We filter out 5 long-tail relations that appear less than 40 times in the entire training set and then classify them into 4 broad categories based on label prefixes in NYT.
As shown in Figure \ref{fig:cluster}, the topology between the relations indicates the semantic connections between them. 
For example, the two types of sport relations almost overlap, while the six location relations (including \textit{country}, \textit{capital}, etc.) are all scattered in the lower left corner.
However, we observe an exception on ``People'', where the 6 subcategories are not spread together spatially. 
We attribute this to semantic divergence. The three subcategories (\textit{place\_lived}, \textit{place\_of\_birth} and \textit{place\_of\_death}) semantically express position information, which is distinct from the other subcategories.
These all demonstrate that our method can break the limits of type-independence and well consider the high-order connections between different relation types.
Besides, we conduct a detailed case study to show the connections between queries in Appendix \ref{app:case}.

\begin{figure}[h!]
  \centering
  \includegraphics[width=0.9\linewidth]{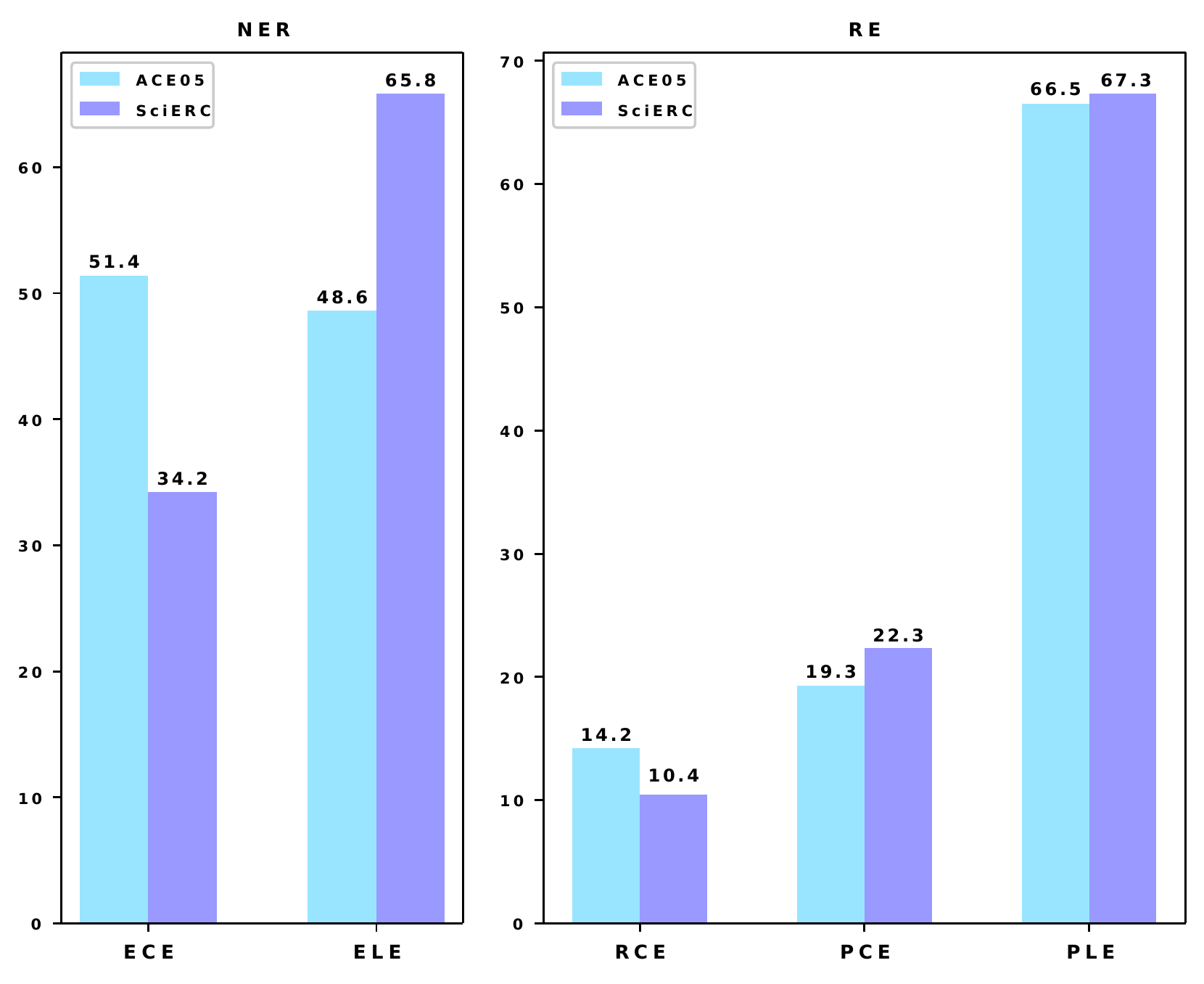}
  \caption{The proportions of different errors in NER and RE on the ACE05 and SciERC test sets.}
  \label{fig:errors}
\end{figure}

\section{Error Analysis}
In this section, we analyze the different error proportions on two tasks: entity classification error (ECE) and entity location error (ELE) for NER, relation classification error (RCE), entity-pair classification error (PCE), and entity-pair location error (PLE) for RE. 
On the NER task, we can observe from Figure \ref{fig:errors} that the proportion of classification errors and localization errors for entity queries on ACE05 is comparable. But on SCiERC, the localization errors are more pronounced. We argue that on account of the longer average length of scientific term entities in SciERC, the identification of entity boundaries becomes more difficult.

As shown in Figure \ref{fig:errors}, identifying the boundary tokens of the subject and object entities is the most difficult on both ACE05 and SciERC. We argue that this is due to the complicated patterns of nesting as well as overlapping in boundary recognition.
In addition, the proportion of relation classification errors is minor on both datasets, which indicates the ability of queries in categorization.

\section{Conclusion}
In this paper, we propose a novel query-based instance discrimination network for relational triple extraction. Based on instance queries, we can extract all types of triples in one step, thus avoiding the problem of error accumulation and relation redundancy. Besides, we set two training objectives for contrastive learning to establish high-order connections between triples while learning rich class-level semantic information. Experimental verifications on five datasets demonstrate that the proposed method reaches state-of-the-art. 

\section*{Limitations}
We discuss here the limitations of the method in this paper.
First, this method requires specifying the relation type in advance and cannot handle unseen categories, which is weak for relation extraction in open domains. 
Second, due to the inherent length limit of BERT, this model cannot deal with excessively long texts. Besides, for the case of a single entity corresponding to multiple mentions in a long document, it can pose a great challenge to the proposed method. 
Finally, incorporating prior knowledge into query embeddings is a promising direction for optimization.

\section*{Acknowledgments}

% This work is supported by the National Key Research and Development Project of China (No. 2018AAA0101900), the Chinese Knowledge Center of Engineering Science and Technology (CKCEST) and MOE Engineering Research Center of Digital Library.
This work is supported by the National Key Research and Development Project of China (No. 2018AAA0101900),  the Key Research and Development Program of Zhejiang Province, China (No. 2021C01013), CKCEST, and MOE Engineering Research Center of Digital Library.

\bibliographystyle{acl_natbib}
\bibliography{anthology}

\clearpage
\newpage

\appendix

\section{Number of queries}
\label{app:queries}
Since the number of queries is pre-fixed, we conduct a comparison experiment in Table \ref{tab:queries} to find the suitable setting. We employ the experiment on SciERC with a maximum number of 10 triples in one sentence. We observe that when the number is obviously larger than the ground truth, the performance of the model does decrease.
We argue that this is due to the imbalance between positive and negative samples, as redundant queries predict the none category $\varnothing$. 
Eventually, the query number $M$ in our experiments is set to 15, which is deployed on all the other datasets.

% Table generated by Excel2LaTeX from sheet 'Sheet4'
\begin{table}[h]
  \centering
  \small
  \setlength\tabcolsep{2.5mm}
  \renewcommand\arraystretch{1.}
    \begin{tabular}{ccccccc}
    \toprule
    \multirow{2}[4]{*}{$M$} & \multicolumn{3}{c}{NER} & \multicolumn{3}{c}{RE} \\
\cmidrule(lr){2-4} \cmidrule(lr){5-7}          & \multicolumn{1}{c}{P} & \multicolumn{1}{c}{R} & \multicolumn{1}{c}{F} & P     & R     & F \\
    \midrule
    10 & \textbf{68.2} & 70.0  & 69.1  & \textbf{40.1}  & 37.8  & 38.9  \\
    15 & 68.1  & \textbf{71.2}  & \textbf{69.6}  & 39.7  & \textbf{38.9}  & \textbf{39.3}  \\
    20 & 67.1  & 70.6  & 68.8  & 39.1  & 37.7  & 38.4  \\
    30 & 66.9  & 70.2  & 68.5  & 38.2  & 36.8  & 37.5  \\
    50 & 68.1  & 70.3  & 69.2  & 38.6  & 36.1  & 37.3  \\
    100 & 66.2 & 70.5  & 68.3  & 39.9  & 35.9  & 37.8  \\
    \bottomrule
    \end{tabular}%
  \caption{The performance impact of different number of queries on the SciERC development set.}
  \label{tab:queries}%
\end{table}%

% Table generated by Excel2LaTeX from sheet 'Sheet4'
\begin{table}[h!]
  \centering
  \small
  \renewcommand\arraystretch{1.}
    \begin{tabular}{llllccc}
    \toprule
    \multirow{2}[0]{*}{Inter-task} & \multicolumn{3}{c}{NER} & \multicolumn{3}{c}{RE} \\
\cmidrule(lr){2-4} \cmidrule(lr){5-7}          & \multicolumn{1}{c}{P} & \multicolumn{1}{c}{R} & \multicolumn{1}{c}{F} & P     & R     & F \\
    \midrule
    Default & \textbf{68.1}  & \textbf{71.2}  & \textbf{69.6}  & \textbf{39.7}  & \textbf{38.9}  & \textbf{39.3} \\
    w/o ent-rel & 67.6  & 70.9  & 69.2  & 38.5  & 37.9  & 38.2  \\
    w/o rel-ent & 67.5  & 70.4  & 68.9  & 39.5  & 38.3  & 38.9  \\
    w/o both & 67.2  & 69.8  & 68.5  & 38.3  & 37.5  & 37.9  \\
    \bottomrule
    \end{tabular}%
  \caption{The performance with different attention mask settings on the SciERC development set.}
  \label{tab:mask}%
\end{table}%

\section{Analysis of query branches}
\label{app:analysis}
\begin{figure}[t!]
  \centering
  \includegraphics[width=0.65\linewidth]{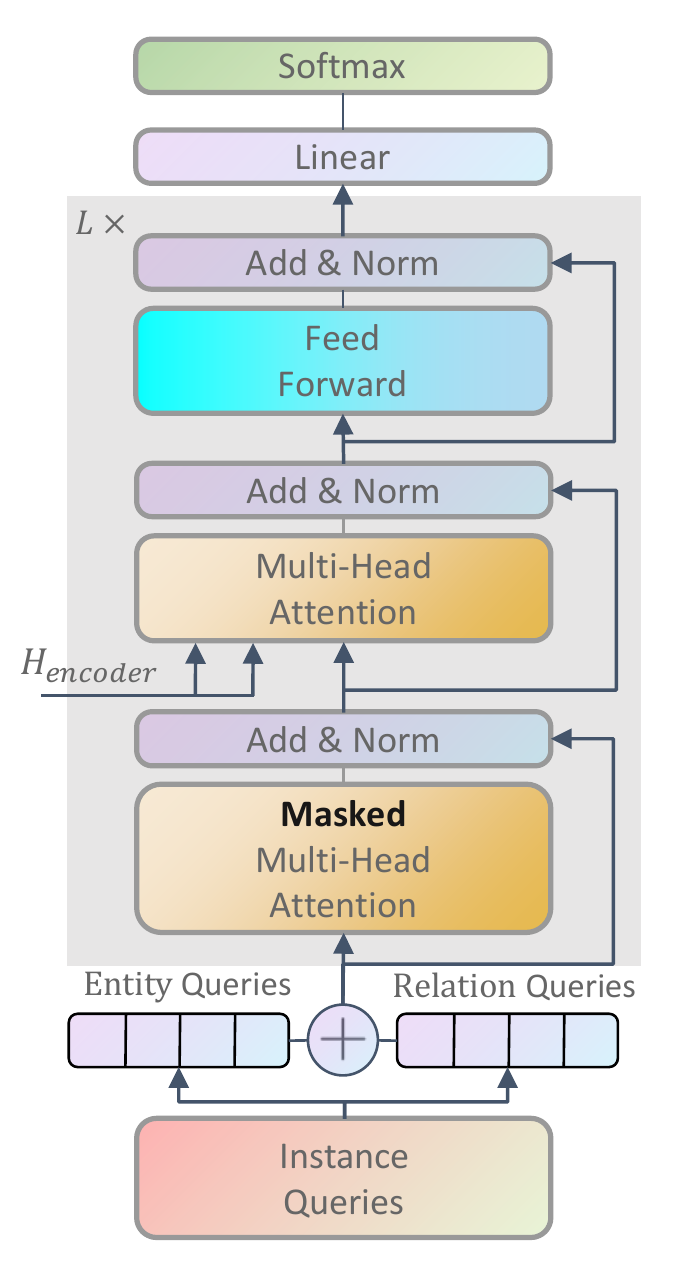}
  \caption{The entity and relation query branches with masked attention in decoder.}
  \label{fig:branch}
\end{figure}

In this section, we explore the entity and relation query branches in our decoder, the detailed architecture is shown in Figure \ref{fig:branch}.
Based on different attention mask matrices for queries, there are three settings.
(1) \textbf{w/o ent-rel}: entity queries are not visible to relation queries.
(2) \textbf{w/o rel-ent}: relation queries are not visible to entity queries.
(3) \textbf{w/o both}: entity queries and relation queries are not visible to each other.
As shown in Table \ref{tab:mask}, if entity queries are visible to relation queries, the performance of the model on RE will improve by +1.1\% (from 38.2\% to 39.3\%). Correspondingly, if relation queries are visible to entity queries, the model performance on NER will increase by +0.7\% (from 68.9\% to 69.6\%).
Further, if they are visible to each other, the F1-measure of the model on NER and RE will be improved by +1.1\% and +1.4\%, respectively. We attribute this to the fact that we can model the inherent dependencies between NER and RE tasks through the attention between entity queries and relation queries.

\begin{figure*}[h!]
  \centering
  \includegraphics[width=\linewidth]{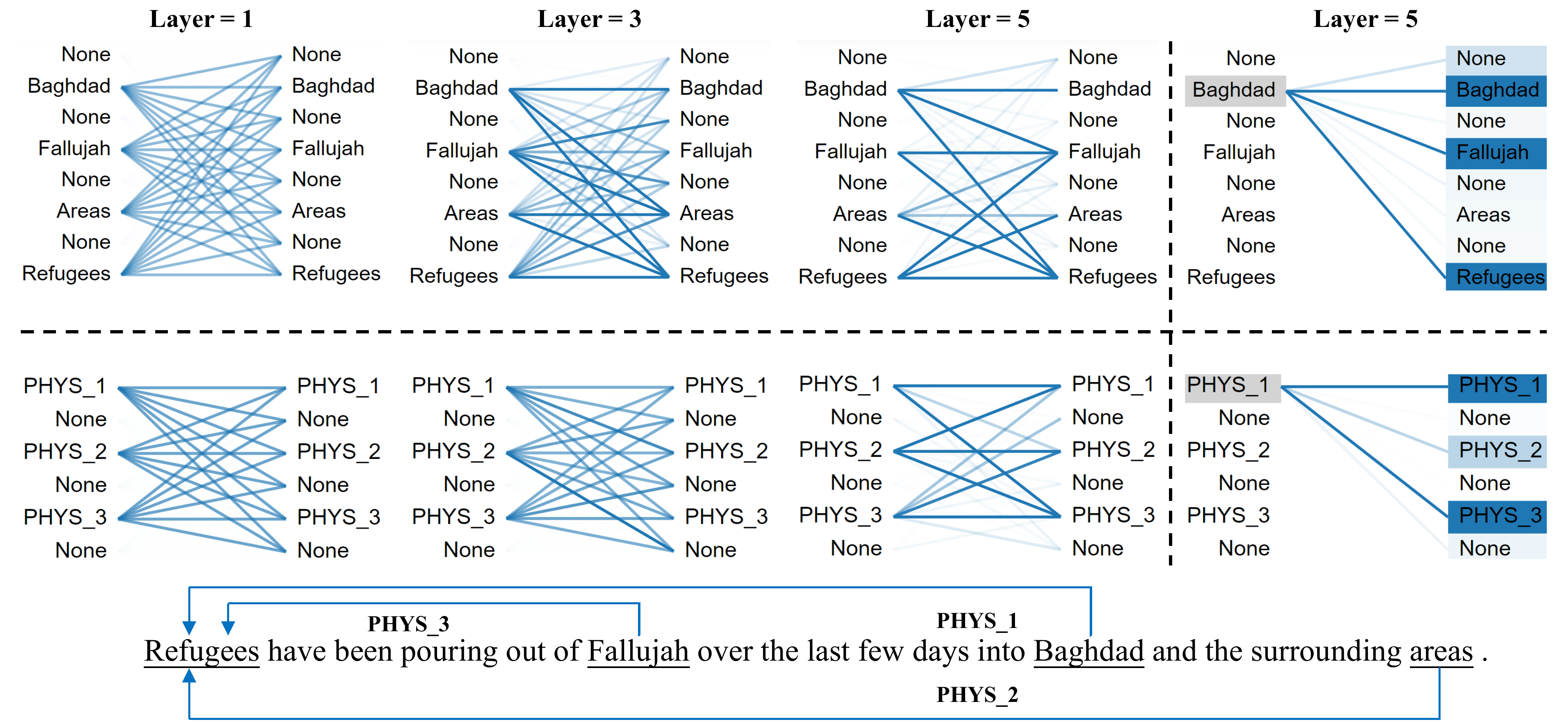}
  \caption{Visualization for the attention weights between entity queries and between relation queries. We randomly select ``None'' with the same number of gold labels, which indicates a non-entity or a non-relation corresponding to a query, and the attention of ``None'' is ignored.
  The sentence is randomly selected from the ACE05 corpus.}
  \label{fig:attn}
\end{figure*}

\begin{table}[t!]
	\centering
	\small
	\setlength{\tabcolsep}{2.5mm}
	\renewcommand\arraystretch{1.}
	\begin{tabular}{@{}lcccccc@{}}
		\toprule
		Dateset & Normal & SEO   & EPO   & SOO \\ 
		\midrule
		NYT  & 3,266  & 1,297 & 978   & 45  \\
		WebNLG  & 239    & 448   & 6     & 85  \\ 
		NYT$^*$  & 3,222  & 1,273 & 969   & 117 \\
		\bottomrule
	\end{tabular}
	\caption{Statistics on the test sets with different triple overlapping patterns.}
	\label{tab:pattern}
\end{table}

\begin{table}[t!]
	\centering
	\small
	\setlength{\tabcolsep}{2.5mm}
	\renewcommand\arraystretch{1.}
	\begin{tabular}{@{}lccccc@{}}
		\toprule
		Dateset  & N=1   & N=2 & N=3 & N=4 & N\textgreater{}5 \\ 
		\midrule
		NYT      & 3,244 & 1,045 & 312   & 291  &108   \\
		WebNLG   & 256   & 175   & 138   & 93   & 41   \\ 
		NYT$^*$  & 3,240 & 1,047 & 314   & 290  &109   \\
		\bottomrule
	\end{tabular}
	\caption{Statistics on the test sets with different triple numbers in each sentence.}
	\label{tab:triple}
\end{table}

\section{Case study}
\label{app:case}
In order to provide a more intuitive understanding of the connection between queries, we visualize the attention weights between entity queries and between relation queries in the decoder with bertviz \citep{vig-2019-multiscale}. 
First, we analyze the connection between the queries corresponding to the gold labels and the queries corresponding to the ``None'' labels, and then we analyze the connection between queries corresponding to the gold labels.

% As shown in Figure \ref{fig:attn},
As shown in the left three columns in Figure \ref{fig:attn},
the attentions are weighted equally between queries  before entering the decoding layers. 
However, as the decoding layer deepens, the queries corresponding to the gold labels will gradually neglect the none labels. 
For entity queries, when the number of decoding layers is 5, the entity queries corresponding to gold entities have most of their attention focused on themselves. 
The same holds true for relation queries.
The observation suggests that these queries do capture dependencies between entities and between relations.  And they are capable of considering the gold and none labels separately during decoding.

Beyond being able to distinguish between the gold and none labels, the queries corresponding to the gold labels have different attention weights from each other.
From the right column of Figure \ref{fig:attn}, we can notice that the entity ``Baghdad'' pays more attention to ``Refugees'' and ``Fallujah'' but ignores ``Areas'' which is a \texttt{LOC} entity. This is due to the fact that ``Baghdad'' has the relation \texttt{PHYS\_1} with ``Refugees'', and has the same entity type \texttt{GPE} as ``Fallujah''.
Similarly, the relation \texttt{PHYS\_1} shows more attention to \texttt{PHYS\_3} than to \texttt{PHYS\_2}, since \texttt{PHYS\_1} and \texttt{PHYS\_3} have the same subject entity ``Refugees'' and the same type \texttt{GPE} of object entity.
It indicates that our approach can well exploit the dependencies between triples.

\end{document}